\documentclass[runningheads]{llncs}

\usepackage{graphicx}
\usepackage{marvosym}
\usepackage{caption}
\usepackage{subcaption}
\usepackage{multirow}
\usepackage{multicol}
\usepackage{amsmath}
\usepackage{xurl}
\begin{document}
\title{From strings to data science: a practical framework for automated string handling}
\titlerunning{From strings to data science}
%
\vspace{-1mm}

\author{John W. van Lith (\Letter) \and
Joaquin Vanschoren}
\authorrunning{J. W. van Lith et al.}
%
\institute{Faculty of Mathematics and Computer Science, Eindhoven University of Technology \\
\email{jlith1997@gmail.com}, \email{j.vanschoren@tue.nl}}
\maketitle              
\vspace{-5mm}

\begin{abstract}
Many machine learning libraries require that string features be converted to a numerical representation for the models to work as intended. Categorical string features can represent a wide variety of data (e.g., zip codes, names, marital status), and are notoriously difficult to preprocess automatically. In this paper, we propose a framework to do so based on best practices, domain knowledge, and novel techniques. It automatically identifies different types of string features, processes them accordingly, and encodes them into numerical representations. We also provide an open source Python implementation\footnote{Open-source library: \url{https://github.com/ml-tue/automated-string-cleaning}} to automatically preprocess categorical string data in tabular datasets and demonstrate promising results on a wide range of datasets.
\end{abstract}
%
\section{Introduction}
\vspace{-2mm}
    Datasets acquired from the real world often contain categorical string data, such as zip codes, names, or occupations. Many machine learning algorithms require that such string features be converted to a numerical representation to work as intended. Depending on the type of data, specific processing is required. For example, geographical string data (e.g., addresses) may be best expressed by latitudes and longitudes. Data scientists are required to manually preprocess such unrefined data, requiring a significant amount of time, up to 60\% of their day \cite{crowdflower2016}. Automated data cleaning tools exist but often fail to robustly address the wide variety of categorical string data. This paper presents a framework that systematically identifies various types of categorical string features in tabular datasets and encodes them appropriately. We also present an open-source Python implementation that we evaluate on a wide range of datasets.

\vspace{-1mm}
\section{Challenges and Related Work}
\vspace{-3mm}
    Our framework addresses a range of challenges. First, \emph{type detection} aims to identify predefined `types' of string data (e.g., dates) that require special preprocessing. Probabilistic Finite-State Machines (PFSMs) \cite{ceritli2020ptype} are a practical solution based on regular expressions and can produce type probabilities. They can also detect missing or anomalous values, such as numeric values in a string column.

    \emph{Statistical type inference} predicts a feature's statistical type (e.g., ordinal or categorical) based on the intrinsic data distribution. Valera et al. \cite{valera2017stattype} use a Bayesian approach to discover whether features are ordinal, categorical, or real-valued, although some manual assessment is still needed. Other techniques that predict classes using features of the data use random forest and gradient boosting classifiers \cite{grabczewski2005feature}, which could be leveraged to achieve class prediction for the statistical type.
    
    \emph{Encoding techniques} convert categorical string data to numeric values, which is challenging because there may be small errors (e.g., typos) and intrinsic meaning (e.g., a time or location). Cerda et al. \cite{cerda2020encoding,cerda2018similarity} use string similarity metrics and min-hashing to tackle morphological variants of the same string. Geocoding APIs (e.g., pgeocode and geopy) can convert geographical strings to coordinates \cite{geopy,pgeocode}. For ordinal string data, heuristic approaches exist that could determine order based on antonyms, superlatives, and quantifiers, e.g. using WordNet \cite{miller1995wordnet} or sentiment intensity analyzers (e.g. VADER, TextBlob, and FlairNLP) \cite{akbik2019flair,hutto2014vader,loria2018textblob}.
    
    Methods have been proposed that recognize, categorize, and process different string entities based on regular expressions \cite{shahbaz2012automated} or domain knowledge that can outperform human experts \cite{hardesty2015,kanter2015deep}. Data cleaning tools exist that are manually operated \cite{dataladder,datacleaner,openrefine2020}, semi-automated \cite{musleh2020coclean,zoller2019benchmark}, or fully automated \cite{krishnan2015sampleclean,krishnan2016activeclean,alphaclean,mahdavi2019raha,holoclean}. At present, however, these automated tools do not focus on string handling \cite{krishnan2015sampleclean,krishnan2016activeclean,holoclean} or they focus on specific steps such as error correction \cite{alphaclean,mahdavi2019raha}.

\begin{figure}[t]
    \includegraphics[width=\textwidth]{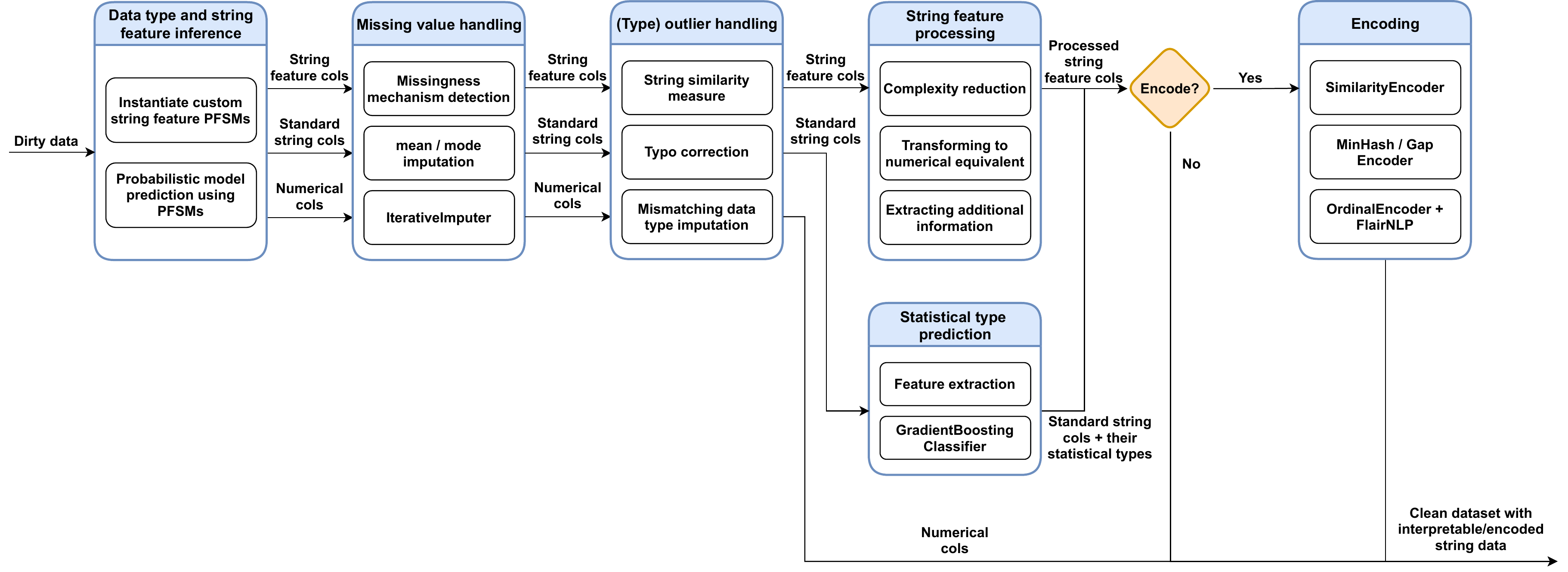}
    \caption{The overall workflow of the framework.} 
    \label{fig:workflow-gen}
\vspace{-4mm}
\end{figure}

\vspace{-3mm}

\section{Methodology}
    Our framework, shown in Fig. \ref{fig:workflow-gen}, is designed to detect and appropriately encode different types of string data in tabular datasets. First, we use PFSMs to infer whether a column is numerical, a known type of string feature (e.g., a date), or any other type of `standard' string data. Based on this first categorization, appropriate missing value and outlier handling methods are applied to the entire dataset to repair inconsistencies. Next, columns with recognized string types go through intermediate type-specific processing, while the remaining columns are classified based on their statistical type (e.g., nominal or ordinal). Finally, the data is encoded by applying the most fitting encoding for each feature.
    
    \subsection{String feature inference}\vspace{-2mm}
        In the first step, we build on PFSMs and the ptype library \cite{ceritli2020ptype}. We created PFSMs based on regular expressions for nine types of string features: coordinates, days, e-mail addresses, filepaths, months, numerical strings, sentences, URLs, and zip codes. A detailed description for each of these can be found in Appendix \ref{app:details}.
        
    \vspace{-3mm}
    \subsection{Handling missing values and outliers}\vspace{-2mm}
        Next, missing values are imputed based on the missingness of the data according to Little's test \cite{little1988test,rubin1976inference} (missing at random, missing not at random, missing completely at random) using mean/mode imputation or a multivariate imputation technique \cite{scikit-learn}. Minor typos are corrected using string metrics \cite{levenshtein1966binary}, and data type outliers are corrected if applicable to ensure robustness in the remaining steps.

    \vspace{-3mm}
    \subsection{Processing inferred string features}\vspace{-2mm}
        Next, we perform intermediate processing of all the string types identified by the PFSMs. First, we simplify the strings, for instance, by only taking the nouns in sentences. Second, we assign or perform specific encoding techniques, such as replacing a date with year-month-day values. Third, we include additional information, such as fetching latitude and longitude values for zip codes. 

        \begin{figure}[t]
            \centering
            \includegraphics[width=0.9\textwidth]{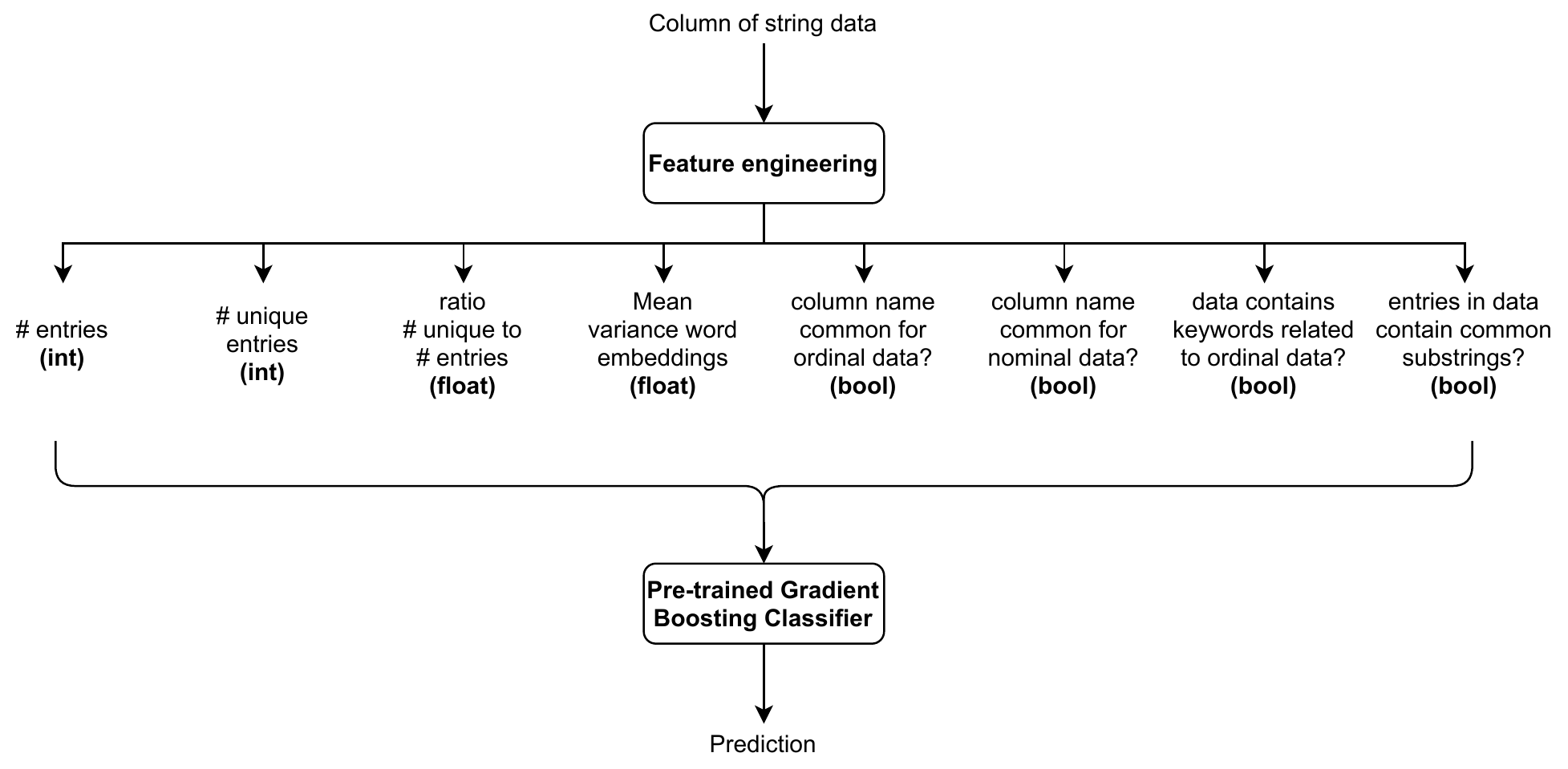}
            \caption{Workflow to predict the statistical type for a given string column.}
            \label{fig:gbc-overview}
        \vspace{-5mm}
        \end{figure}
    
    \vspace{-3mm}
    \subsection{Statistical type prediction}\vspace{-2mm}
        String features not identified by the PFSMs are marked as `standard' strings. For these, we infer their statistical type, i.e., whether they contain ordered (ordinal) data or unordered (nominal) data. The prediction is based on eight properties extracted from the feature, shown in Fig. \ref{fig:gbc-overview}, including the uniqueness of the string values, whether the column name or values suggests ordinality, and whether a GloVe word embedding of the string values shows clear relationships between the values. The rationale behind these is explained in Appendix \ref{app:details2}. These features are fed to a gradient boosting classifier to predict the statistical type. This classifier was trained on real-world features, manually annotated, listed in Appendix \ref{app:datasets}.

    \subsection{Encoding}
        Finally, an appropriate encoding is applied for each categorical string feature. For the string types identified by the PFSMs, a predefined encoding is applied, per Appendix \ref{app:details}. For nominal data, we use category encoders and target encoding due to their robustness to morphological variants and high cardinality features \cite{cerda2020encoding,cerda2018similarity}. We apply the similarity, Gamma-Poisson, and min-hash encoders when the cardinality is below 30, below 100, and at least 100, respectively. For ordinal data, the ordering is defined by a text sentiment intensity analyzer (FlairNLP \cite{akbik2019flair}) that converts string values (e.g., very bad, very good) to an ordered encoding.
        

    \begin{figure}[t]
    \vspace{-4mm}
    \centering
        \includegraphics[width=0.88\textwidth]{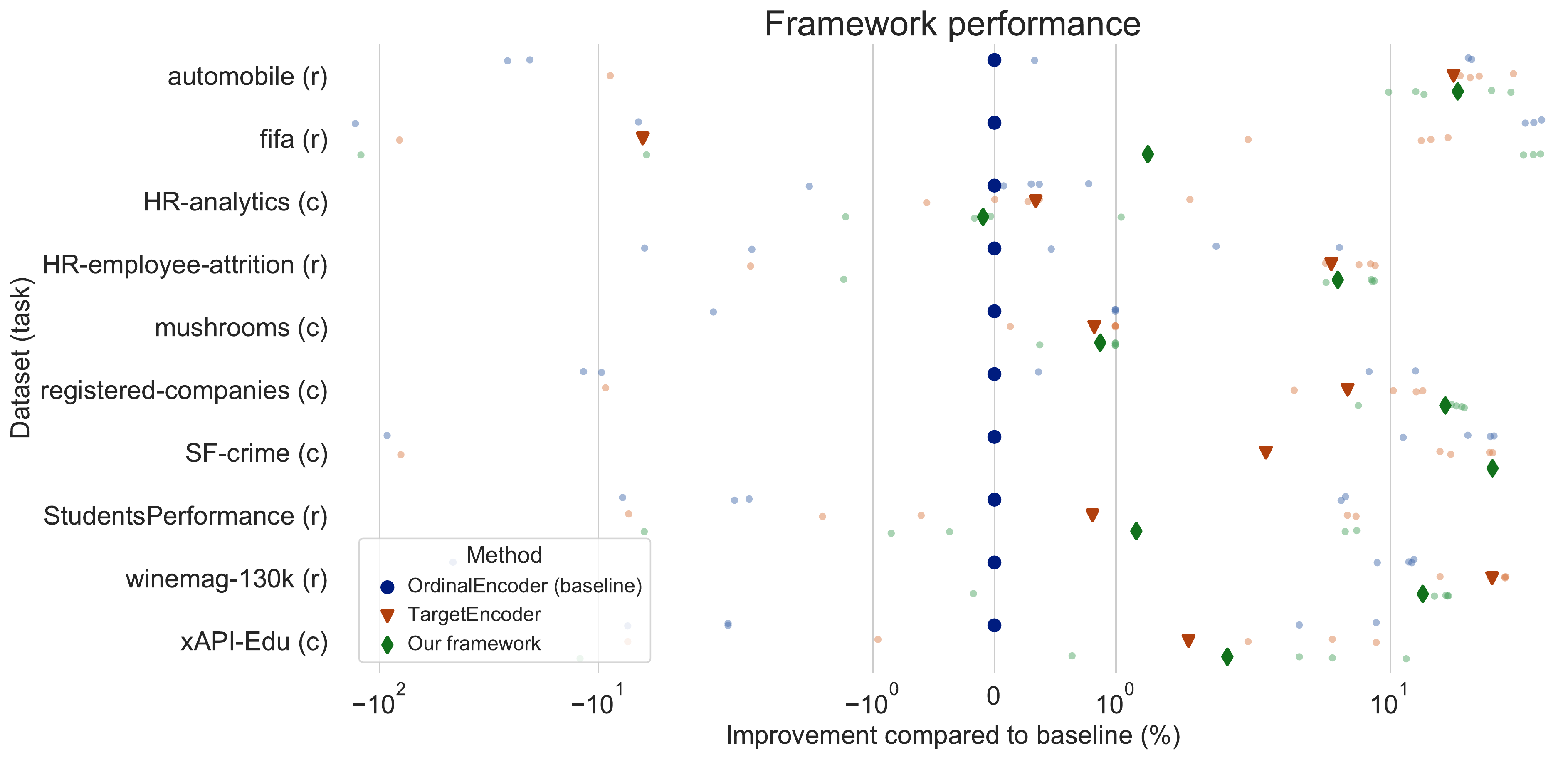}
        \vspace{-3mm}
        \caption{Relative mean performance of our framework against baseline preprocessing. The smaller dots represent the scores of individual folds of each method.}
        \label{fig:fw-perf}
    \vspace{-7mm}
    \end{figure}

\vspace{-3mm}
\section{Evaluation}
    \vspace{-3mm}
    We evaluate our framework and its individual components on a range of real-world datasets with categorical string features, listed in Appendix \ref{app:datasets}. We evaluate the downstream performance of gradient boosting models trained on the encoded data. Boosting models are intrinsically robust against high-dimensional data, thus allowing a fairer comparison. We use stratified 5-fold cross-validation, and measure accuracy for classification tasks and MAE for regression tasks.
    
    \textbf{Global framework evaluation.}
    First, we evaluate the framework as a whole and compare it to a baseline where the data is manually preprocessed using mean/mode imputation and ordinal or target encoding for the categorical string features. Fig. \ref{fig:fw-perf} shows the relative performance differences for each of the five folds and their mean. These results indicate that our framework performs well on real-world data. On the winemag-130k dataset, the automated encoding proves suboptimal, which is likely tied to the specific heuristics used.
    
   \textbf{Feature type inference.}
    In Table \ref{tab:eval-pfsm}, we compare the predictions of our PFSMs against the ground truth feature types. Most PFSMs report perfect accuracy. Filepaths and sentences are detected with 70-80\% accuracy, since the exact format of such data can be unexpected. Outliers are also detected correctly, except for sentence PFSMs, where out of the 130217 entries, 42158 entries were false negatives, which is certainly a point for improvement.
    
    \begin{table}[t]
        \centering
        \resizebox{0.7\textwidth}{!}{%
        \begin{tabular}{|l|l|l|l|l|l|}
        \hline
        String feature & \begin{tabular}[c]{@{}l@{}}Nr. of\\ columns\end{tabular} & \begin{tabular}[c]{@{}l@{}}Nr. of correctly\\ inferred columns\end{tabular} & Accuracy & \begin{tabular}[c]{@{}l@{}}Nr. of false\\ negative outliers\end{tabular} & \begin{tabular}[c]{@{}l@{}}Ratio of false \\ negative outliers\end{tabular} \\
        \hline \hline
        Coordinate     & 2  & 2  & 1.0   & 0      & -           \\
        Day            & 1  & 1  & 1.0   & 0      & -           \\
        E-mail         & 4  & 4  & 1.0   & 0      & -           \\
        Filepath       & 5  & 4  & 0.80  & 0      & -           \\
        Month          & 3  & 3  & 1.0   & 0      & -           \\
        Numerical      & 6  & 6  & 1.0   & 0      & -           \\
        Sentence       & 4  & 3  & 0.75  & 42158  & 0.32        \\
        URL            & 4  & 4  & 1.0   & 0      & -           \\
        Zip code       & 3  & 3  & 1.0   & 0      & -           \\ 
        \hline
        \end{tabular}
        }
        \caption{Results of string feature inference using PFSMs.}
        \label{tab:eval-pfsm}
    \vspace{-5mm}
    \end{table}
    
    \begin{figure}[t]
    \vspace{-2mm}
    \centering
        \includegraphics[width=0.88\textwidth]{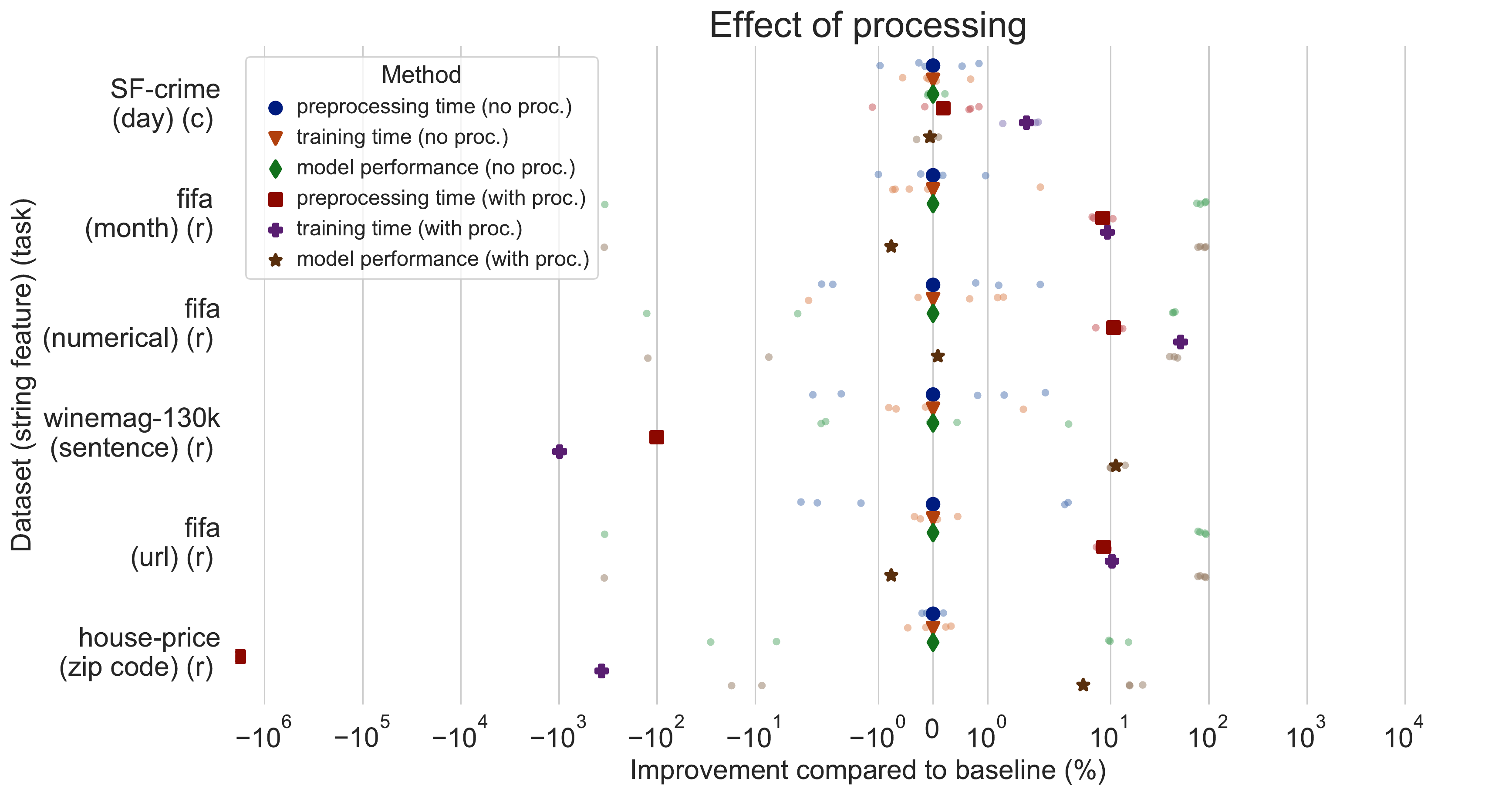}
        \vspace{-2mm}
        \caption{Difference in performance metrics (mean accuracy/MAE, preprocessing time, and training time) with and without processing of known string feature types. The smaller dots represent individual folds or runs.}
        \label{fig:proc-perf}
    \vspace{-7mm}
    \end{figure}
    
    \textbf{Processing inferred string features.}
    Fig. \ref{fig:proc-perf} compares the performance of our framework with and without intermediate processing for the string feature types identified by the PFSMs. For some features, this processing causes a 10\% performance improvement, while on others, it remains about the same. This processing does require extra processing time, caused by API latency and text processing, yet it seems worth the extra time for zip codes and sentences. 
    Moreover, the reduced string complexity (removing redundant words) and conversion of numerical strings (e.g. `$>10$') into numerical representations reduce the training time by at least ten percent on half of the datasets.
    
    \textbf{Statistical type prediction.} Tables \ref{tab:eval-gbc1} and \ref{tab:eval-gbc2} evaluate the gradient boosting classifier that predicts whether standard string features are nominal or ordinal, by comparison against the ground truth using leave-one-out cross-validation. These predictions are highly accurate, with very few misclassifications. Hence, our eight extracted features are highly indicative of ordinality in the feature values.
    
    \begin{table}[t]
    \vspace{-2mm}
    \begin{subtable}[t]{0.25\textwidth}
        \centering
        \resizebox{0.85\textwidth}{!}{%
        \begin{tabular}{|l|l|}
        \hline
        Metric    & Score         \\ 
        \hline \hline
        Accuracy  & 0.980 ± 0.14 \\
        F1 score  & 0.978         \\
        Precision & 0.971         \\
        Recall    & 0.985         \\
        AUC score & 0.980         \\ 
        \hline
        \end{tabular}
        }
        \vspace{-1mm}
        \caption{\small Statistical type prediction scores}
        \label{tab:eval-gbc1}
    \end{subtable}
    \hfill
    \begin{subtable}[t]{0.25\textwidth}
        \centering
        \resizebox{0.95\textwidth}{!}{%
        \begin{tabular}{clcc}
        \multicolumn{2}{l}{\multirow{2}{*}{}}                                                              & \multicolumn{2}{c}{Predicted class}                                    \\ \cline{3-4} 
        \multicolumn{2}{l}{}                                                                               & \multicolumn{1}{|l}{Ordinal} & \multicolumn{1}{l|}{Nominal}             \\ \cline{2-4} 
        \multicolumn{1}{c|}{\multirow{4}{*}{\rotatebox[origin=c]{90}{Actual class}}} & \multicolumn{1}{l|}{\multirow{2}{*}{Ordinal}} & \multirow{2}{*}{79}         & \multicolumn{1}{c|}{\multirow{2}{*}{2}}  \\
        \multicolumn{1}{c|}{}                              & \multicolumn{1}{l|}{}                         &                             & \multicolumn{1}{c|}{}                    \\
        \multicolumn{1}{c|}{}                              & \multicolumn{1}{l|}{\multirow{2}{*}{Nominal}} & \multirow{2}{*}{1}          & \multicolumn{1}{c|}{\multirow{2}{*}{67}} \\
        \multicolumn{1}{c|}{}                              & \multicolumn{1}{l|}{}                         &                             & \multicolumn{1}{c|}{}                    \\ \cline{2-4} 
        \end{tabular}
        }
        \vspace{-0.25mm}
        \caption{\small Statistical type prediction: Confusion matrix}
        \label{tab:eval-gbc2}
    \end{subtable}
    \hfill
    \begin{subtable}[t]{0.32\textwidth}
        \centering
        \resizebox{0.95\textwidth}{!}{%
        \begin{tabular}{|l|ll|}
        \hline
        Ordering method                 & \multicolumn{2}{c|}{Rank Correlation} \\ 
        \hline \hline
        Baseline & 0.1562                & ± 0.5313        \\
        FlairNLP & \textbf{0.7189}       & \textbf{± 0.5356}        \\ \hline
        \end{tabular}
        }
        \vspace{3.4mm}
        \caption{\small Spearman Rank Correlation for ordinal encoders vs. ground truth ordering.}
        \label{tab:eval-enc-order}
    \end{subtable}
    \vspace{-3mm}
    \caption{Results from various modules of the framework.}
    \label{tab:eval-all}
    \vspace{-5mm}
    \end{table}
    
\begin{figure}[t]
    \centering
    \vspace{-2mm}
        \includegraphics[width=0.88\textwidth]{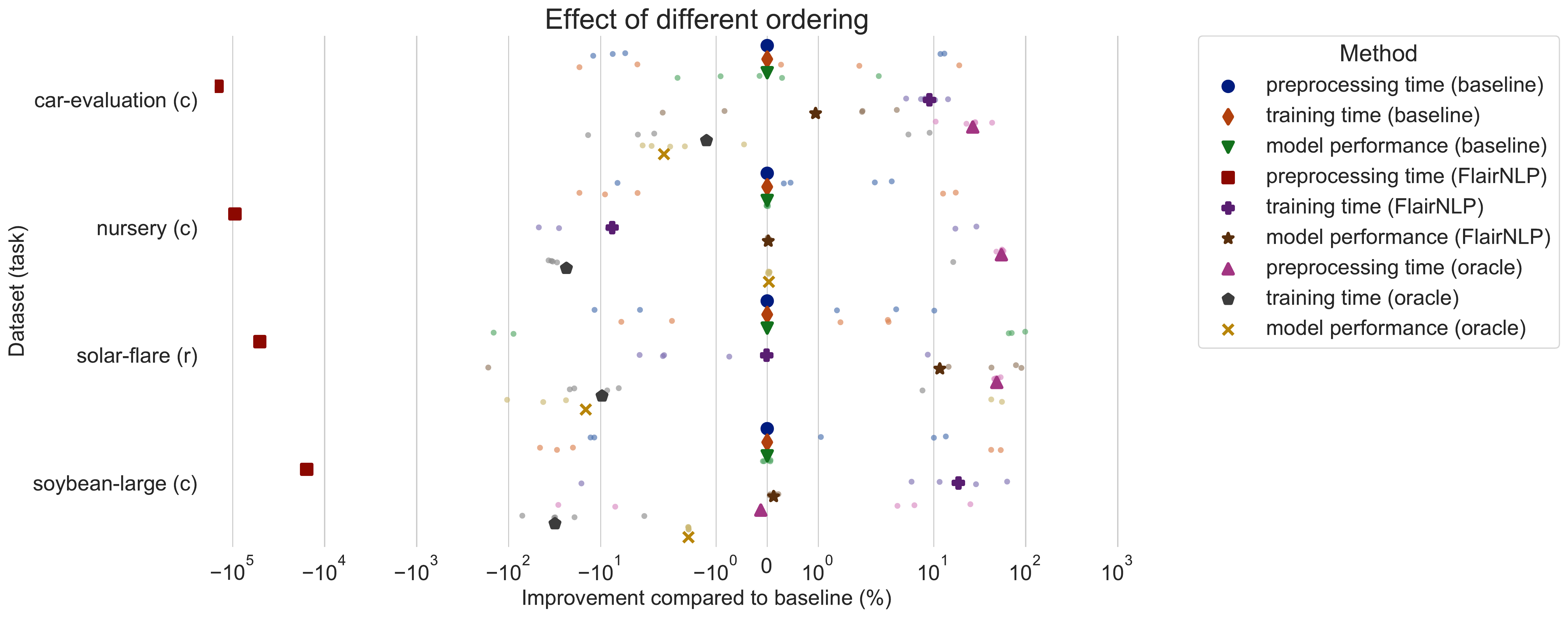}
        \caption{Difference in mean performance metrics between different ordering strategies. The smaller dots represent individual folds or runs.}
        \label{fig:enc-perf}
    \vspace{-6mm}
    \end{figure}

    \textbf{Ordinal encoding.} 
    Finally, we compare the ordinal encoding based on sentiment intensity (FlairNLP) vs. the baseline ordinal encoding by comparing them to an oracle with the ground-truth ordering. Table \ref{tab:eval-enc-order} and Fig. \ref{fig:enc-perf} show that FlairNLP significantly outperforms the baseline, although the effect on downstream model performance is limited ($<1\%$), and FlairNLP does require significantly more preprocessing time.
    
    \vspace{-3mm}
\section{Conclusions and future work}
    \vspace{-4mm}
    The automation of data cleaning is a fledgling open research field. We presented a framework that combines state-of-the-art techniques and additional novel components to enable automated string data cleaning. This framework shows promising results, and some of its novel components (e.g., string feature type inference, ordinality detection, and encoding using FlairNLP) perform very well, especially in terms of identifying special types of categorical string data and adequately processing and encoding them. However, several challenges remain. First, string feature type inference using (regular expression-based) PFSMs is sensitive to the exact formatting of strings. More robust techniques are needed, such as sub-pattern matching and better use of type probabilities in subsequent processing (e.g., if it is only 60\% certain that a string feature represents a date, a more robust encoding is needed, or a human should be brought in the loop). Second, the string type-specific processing was suboptimal or no better than the baseline on some datasets in some aspects, providing interesting cases for further study. Finally, the encoding of ordinal string data still leaves room for improvement. The sentiment intensity-based encoding has shown to perform well on some aspects and poorly on others. We believe that more sophisticated approaches are possible, e.g., paying special attention to numbers appearing in the string data. Overall, we hope that this framework and open-source implementation will speed up research in these areas.
    
    \vspace{-2mm}
    \subsubsection{Acknowledgements} The authors would like to thank Marcos de Paula Bueno on his valuable feedback on this paper. Furthermore, this work was partly funded by the European Research Council under EU Horizon 2020 progamme, project 592215 (TAILOR).

%
%
%
\bibliographystyle{splncs04}
\bibliography{main}

\appendix
\newpage 
\section{Details on string features and processing}\label{app:details}
    \subsubsection{Coordinate}
        This is a string feature that represent GPS or Degree-Minute-Second (DMS) coordinates such as \texttt{N29.10.56 W90.00.00}, \texttt{N29:10:56}, and \texttt{29°} \texttt{10'56.22"N}. Coordinates can be distinguished from other string features based on the following characteristics:
        
        \begin{itemize}
            \item Two sequences of at most two digits and a third sequence which is a float with at most two digits before the decimal point.
            \item A character that separates the three sequences of digits (e.g., \texttt{.} or \texttt{:}). In the context of DMS coordinates, these characters are \texttt{°}, \texttt{'}, and \texttt{"} respectively.
            \item A cardinal direction at the beginning or at the end of the string (i.e., \texttt{N}, \texttt{E}, \texttt{S}, and \texttt{W}).
        \end{itemize}
        
        This string feature is processed as follows. First, the string feature is split up into two separate parts for each coordinate in the entry, in which one part represents the cardinal direction of the coordinate and the other part represents the numerical information. Second, the current format of the coordinate string feature is converted into the corresponding decimal latlong value. The string feature is formatted using degrees, minutes, and seconds. This format can be converted to representative decimal values using the following formula\footnote{Taken from ``Geographic coordinate conversion'' at \url{https://en.wikipedia.org/wiki/Geographic\_coordinate\_conversion}}:
    
        \begin{equation*}
            \text{decimal} = \begin{cases}
                                 -(\text{degrees} + \frac{\text{minutes}}{60} + \frac{\text{seconds}}{3600}) & \text{if } c \in \{ S, W\} \\
                                 \text{degrees} + \frac{\text{minutes}}{60} + \frac{\text{seconds}}{3600} & \text{otherwise}
                             \end{cases}
        \end{equation*}
    
        Last, additional information is extracted in case the string feature contains both the latitude and the longitude values. The additional information that can be extracted includes Earth-Centered Earth-Fixed representations of the latlong value and postal codes and country codes via geopy \cite{geopy}. If the user decides to encode the data, the extracted postal and country codes will receive a nominal encoding in the final step of the framework.
        
    \subsubsection{Day}
        This string feature represent the names of the seven days in the week such as \texttt{Monday}. These names can appear in data in several formats. For example, \texttt{Monday} can be written as \texttt{Mon} and \texttt{Mo}. Days can be distinguished from other string features based on the following characteristics:
        
        \begin{itemize}
            \item A prefix of at least two characters, indicating the day of the week (e.g., \texttt{Mo} for Monday, \texttt{Th} for Thursday, etc.).
            \item The suffix \texttt{day}, if present.
            \item A distinct set of characters that comes after the prefix and before the suffix  (e.g., if  the string is \texttt{Thursday}, then \texttt{Th} should be followed by \texttt{urs}).
        \end{itemize}
        
        As this string feature is the least complex out of all inferred string features, it is also the most simple to process. Considering only the first two characters for days is the most reduction that can be done while still being able to make a distinction between each unique day of the week. If the user decides to encode the data, this string feature will receive a nominal encoding in the final step of the framework.
    
    \subsubsection{E-mail}
        This string feature represents all valid e-mail addresses from any domain such as \texttt{Jane@tue.nl} and \texttt{john.doe@hotmail.co.uk}. This feature can be distinguished from others based on the following characteristics:
        
        \begin{itemize}
            \item The character \texttt{@} which is between two sets of characters.
            \item A substring in front of the \texttt{@} (i.e., the name of the e-mail) which is composed of valid characters (e.g., the e-mail address \texttt{\#@*\%\#\$@hotmail.com} is invalid as the characters before the final \texttt{@} cannot be included in an e-mail name).
            \item A substring that comes after the \texttt{@} which is composed of valid characters and at least one dot inbetween those characters (e.g., \texttt{name@hotmail} is not a valid e-mail address as the domain name is incomplete).
        \end{itemize}
        
        This string feature is processed as follows. We first remove the longest common suffix of all entries. Then, additional special characters are removed to simplify the values. If the user decides to encode the data, this string feature will receive a nominal encoding in the final step of the framework.
    
    \subsubsection{Filepath}
        This string feature represents paths within a local system such as \texttt{C:/Windows/} and \texttt{C:/Users/Documents}. Filepaths can be distinguished from other string features by the following characteristics:
        
        \begin{itemize}
            \item A series of substrings which are separated from each other using either \texttt{/} or \texttt{\char`\\} (e.g., \texttt{home/users}). 
            \item Each substring cannot contain any of the following characters: \texttt{\char`\\/:*?"<>|}
            \item If present, a prefix that represents the root disk or a sequence of dots followed by a slash or a backslash (e.g., \texttt{C:/}, \texttt{../}, etc.).
        \end{itemize}
        
        Processing this string feature is similar to how e-mail addresses are processed and is aimed to reduce string complexity. For this feature, the longest common prefix and suffix are removed from all entries and all special characters are removed. If the user decides to encode the data, this string feature will receive a nominal encoding in the final step of the framework.
    
    \subsubsection{Month}
        This string feature represents the non-numerical representation of a month with or without year and day and can be distinguished based on the following criteria:
        
        \begin{itemize}
            \item A prefix of at least three characters, representing a unique month (e.g., \texttt{Apr}).
            \item If present, the remaining substring that comes after the prefix (e.g., \texttt{il} comes after the prefix \texttt{Apr}).
            \item If present, a sequence of at most two digits before or after the month which represents a day in the month (e.g., \texttt{1 January} or \texttt{January 1}). 
            \item If present, a sequence of at most four digits or a sequence with prefix \texttt{'} followed by two digits that comes after the month which represents the year (e.g., \texttt{January 2000} or \texttt{January '00}). Both day and year can be present at the same time.
        \end{itemize}
        
\begin{figure}
            \includegraphics[width=\textwidth]{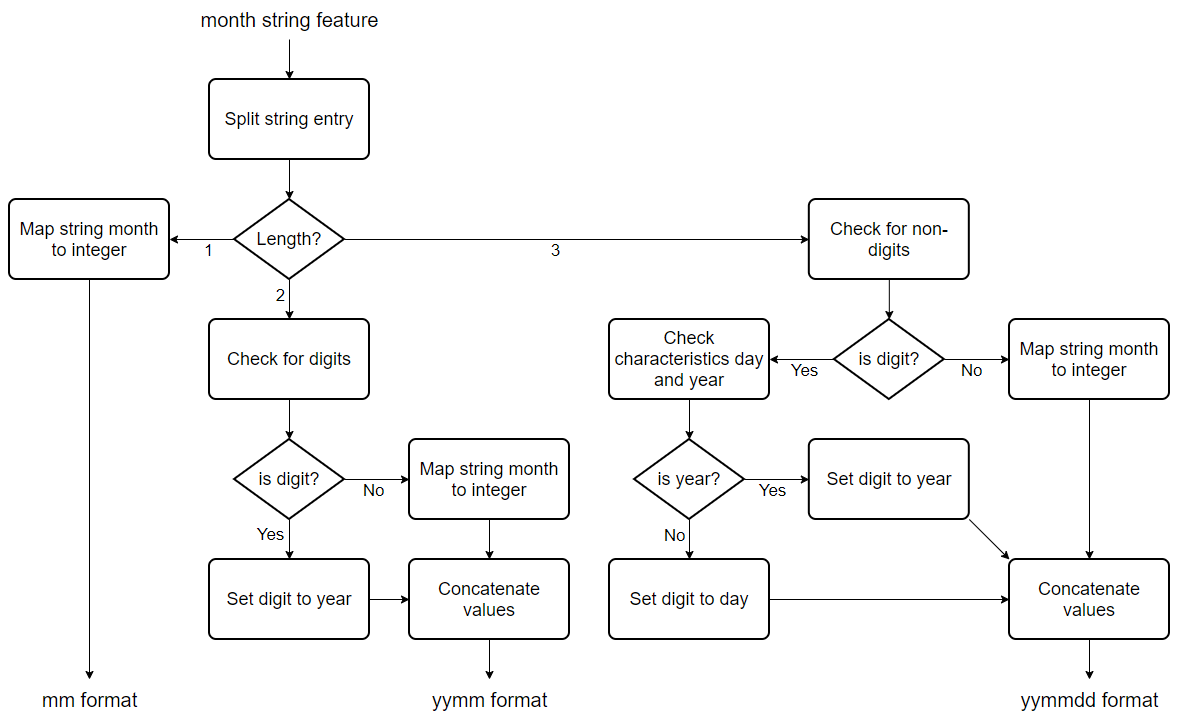}
            \caption{The overall workflow for processing month string features.}
            \label{fig:proc-month}
        \end{figure}

        Processing this string feature is based on the format that is being presented. Each format is split up into individual components that represent either a day, month, or year. The key step in this procedure is to ensure that the string representative of the month is turned into the corresponding numerical representation. After this key step is performed, all values are concatenated to each other according to the format \texttt{yyyymmdd}. The overall workflow of this processing step is depicted in Fig. \ref{fig:proc-month}. As the string feature is already transformed to its numerical representation, no encoding would be required if the user requested so.
        
    \subsubsection{Numerical}
        There are a variety of entries that are relatively easy for users to distinguish as numerical values but are usually inferred as strings by any type detection or inference technique due to the presence of certain non-numerical characters. It is therefore important to categorize such entries as a string feature to properly handle and process them. We obtain numerical string features using any of the following characteristics:
        
        \begin{itemize}
            \item Between two sequences of digits, one of the following characters: \texttt{-+\_/:;\&'} A space or the substring \texttt{to} is also applicable (e.g., \texttt{100 to 200}).
            \item Before a single sequence of digits, any of the following words: \texttt{Less than}, \texttt{Lower than}, \texttt{Under}, \texttt{Below}, \texttt{Greater than}, \texttt{Higher than}, \texttt{Over}, \texttt{Above}.
            \item Before or after a single sequence of digits, any of the following characters: \texttt{<>+\$\%=}
        \end{itemize}
        
        Numerical string features are processed based on what they represent. If the string feature represents a range of values, the mean of the range is calculated for each entry. After that, the range entries are encoded according to the numerical order of the ranges. If the string feature does not represent a range, we remove all special characters and consider all resulting numbers as separate numerical features. As the string feature is already transformed to its numerical representation, no encoding would be required if the user requested so.
    
    \subsubsection{Sentence}
        This string feature is composed of a sequence of words, typically found in datasets that contain reviews or descriptions. It is slightly more difficult to express this string feature as a regular expression compared to the others because of its overlapping characteristics with regular string entries that consists of a couple of words. However, it is still possible to perform string feature inference for sentences based on the following characteristics:
        
        \begin{itemize}
            \item A substring of characters followed by a space for at least five times (i.e., the entry is at least six words long).
        \end{itemize}
    
        The goal for processing sentence string features is to remove redundancy in the entry and to make these more relevant for use in tabular data. The technique used to achieve this goal is the NLTK word tokenizer, which takes a sentence and divides these into tuples containing each word and their associated part of speech. Then, every word that is associated with a noun is joined together with a space into a single string which is then passed on. The result is a group of nouns that are supposed to represent the essence of the sentence and are ready to be encoded in the next step of the library. If the user decides to encode the data, this string feature will receive a nominal encoding in the final step of the framework.

    \subsubsection{URL}
        This string feature represents any link to a website or domain such as \texttt{https://www.tue.nl/} and \texttt{http://canvas.tue.nl/login}. The characteristics of this string feature is similar to that of filepaths, with a few exceptions:
        
        \begin{itemize}
            \item An optional suffix which represents a certain protocol (e.g., \texttt{http://}).
            \item A series of at most four character sequences, separated from each other by a dot (e.g., \texttt{www.google.com} or \texttt{google.com}). Note that the last (pair of) sequence(s) contain(s) at most three characters.
        \end{itemize}
    
        Processing this string feature follows the same procedure as filepaths. If the user decides to encode the data, this string feature will receive a nominal encoding in the final step of the framework.
    
    \subsubsection{Zip code}
        This string feature represents zip or postal codes from a handful of countries. Note that we are only able to infer zip codes that contain non-numerical characters as numerical-only zip codes are much more difficult to infer using PFSMs without overlapping actual numerical features.
        
        Processing this string feature is mainly done to extract additional information from each entry. In our case, we make use of the geopy library to fetch the latitude, longitude, and country code of the zip code. Furthermore, we also calculate the ECEF coordinate using the latitude and longitude values. If the user decides to encode the data, the zip code string feature will receive a nominal encoding in the final step of the framework.

\section{Details on ordinality feature extraction}\label{app:details2}
    \begin{itemize}
            \item \textit{The total number of rows in the column}: It is possible that the number of rows in combination with other extracted features can increase the performance of the classifier. Obtaining this value is done by measuring the length of the column.
            
            \item \textit{The number of unique values in the column}: In general, nominal data tends to vary more in cardinality as opposed to ordinal data. Furthermore, some ordinal data columns tend to adhere to Likert-scale characteristics regarding the possible number of unique entries, which also limits its cardinality. The value is obtained by counting all unique entries in a column.
            
            \item \textit{The ratio between the number of unique values and the total number of rows}: As a rule of thumb, some domain experts tend to classify data as ordinal when the ratio between the unique values and the total number of rows is at most 0.05. The ratio for nominal data tends to be at most 0.2. As a result of this rule of thumb, we extract the ratio for use in the classifier by dividing the number of unique values by the total number of rows.
            
            \item \textit{The mean of the variance of the distance between the word embeddings of unique entries}: The idea behind extracting this feature is that the word embeddings of certain entries showcase interesting linear substructures in the word vector space. By taking a pre-trained word vector space, the classifier may be able to make a distinction between ordered and unordered data based on differences in the substructure. The first step is to split each entry into a set of words which are then embedded using a pre-trained word vector space. This work makes use of the Wikipedia word vector space by GloVe, which consists of over 400 000 words in the vocabulary embedded into a 50-dimensional vector space \cite{pennington2014glove}, to assign each word in an entry to a vector. A random point in the vector space will be assigned to a word in case it does not appear in the pre-trained corpus. Next, the mean of all dimensions for each word vector in the entry is calculated such that all word vectors of the entry are now represented as a single point in the 50-dimensional vector space. After this is done for all entries in the column, the variance between each dimension is calculated. Finally, the mean of each dimension is taken and the resulting value is a single float value that can be used to potentially distinguish ordered data from unordered data.
            
            \item \textit{Whether the column name is commonly used in ordinal data}: There are a set of keywords that can commonly be found in column names for ordinal data. Examples of certain keywords include \texttt{grade}, \texttt{stage}, and \texttt{opinion}. By checking whether a column name is contained within one of those keywords and vice versa, we are able to tell whether the data in the column is more likely to be ordinal or not. Note that keywords for column names used in this work are created based on domain expertise, which means that results may vary when other keywords are used.
            
            \item \textit{Whether the column name is commonly used in nominal data}: The approach of extracting this feature is similar to that of checking ordinal traits in the column name, except that the names obtained via domain expertise are now commonly used in nominal data. Typical nominal column names include \texttt{address}, \texttt{city}, \texttt{name}, and \texttt{type}. Again, since the used names are based on domain expertise, results in performance may vary when other names are used.
            
            \item \textit{Whether the unique entries contain keywords that are commonly found in ordinal data}: Extracting this feature is similar to the two aforementioned techniques, except that ordinality will now be implied based on keywords that are commonly found in ordinal data. The keywords that were used in this work are adjectives and nouns that are typically found in Likert-scale questionnaires.
            
            \item \textit{Whether the unique entries share a number of common substrings}: Ordinal data tends to contain entries with overlapping substrings. For example, the strings \texttt{disagree}, \texttt{agree}, and \texttt{wholeheartedly agree} all contain the substring \texttt{agree}. By checking whether there are common substrings in the data of sufficient length, it is possible that the classifier associates the occurrence of substrings with the implication that the data is ordinal. 
        \end{itemize}

\newpage
\section{Datasets}\label{app:datasets}
    A list of all datasets that were used during the evaluation are provided here, including additional information on what was used from the data.

    \subsection{String feature inference and processing}\label{app:datasets-string}
    Some of the datasets listed below were used to evaluate both string feature inference and string feature processing. The datasets without model, task, and target were only used to evaluate string feature inference based on the manually assigned ground truth of the columns. The datasets with the previously mentioned components were used for both string feature inference (with manual labeling of string feature) and string feature processing (evaluation by running the associated task).

    \subsubsection{Coordinate}
    \begin{itemize}
        \item \textbf{Digital altimetric data information - GPS}. Columns: \textit{latgms, loggms}. \url{https://www.kaggle.com/mpwolke/cusersmarildownloadsgpscsv}
    \end{itemize}
        
    \subsubsection{Day}
    \begin{itemize}
        \item \textbf{San Francisco Crime Classification}. Columns: \textit{DayOfWeek}. Model: \textit{GradientBoostingClassifier}. Task: \textit{Classification}. Target: \textit{Category}. \\ \url{https://www.kaggle.com/kaggle/san-francisco-crime-classification}
    \end{itemize}
        
    \subsubsection{E-mail}
    \begin{itemize}
        \item \textbf{Data\_UK}. Columns: \textit{email}. \url{https://www.kaggle.com/phool1804/data-uk}
        
        \item \textbf{Enrico's Email Flows}. Columns: \textit{sender, receiver}. \\ \url{https://www.kaggle.com/emarock/enricos-email-flows}
        
        \item \textbf{Indian Companies Registration Data [1857 - 2020]}. Columns: \textit{EMAIL \_ADDR}. \url{https://www.kaggle.com/rowhitswami/all-indian-companies-registration-data-1900-2019}
    \end{itemize}
    
    \subsubsection{Filepath}
    \begin{itemize}
        \item \textbf{Collection of Classification \& Regression Datasets}. Columns: \textit{Image Index}. \url{https://www.kaggle.com/balakrishcodes/others?select=xrayfull.csv}
        
        \item \textbf{Hillary Clinton's Emails}. Columns: \textit{MetadataPdfLink}. \\ \url{https://www.kaggle.com/kaggle/hillary-clinton-emails?select=Emails.csv}
        
        \item \textbf{Liver and Liver Tumor Segmentation}. Columns: \textit{filepath, liver\_maskpath, tumor\_maskpath}. \url{https://www.kaggle.com/andrewmvd/lits-png?select=lits\_df.csv}
    \end{itemize}
    
    \subsubsection{Month}
    \begin{itemize}
        \item \textbf{FIFA 19 complete player dataset}. Model: \textit{GradientBoostingRegressor}. Task: \textit{Regression}. Target: \textit{Value}. Columns: \textit{Joined}. \\ \url{https://www.kaggle.com/karangadiya/fifa19}
        
        \item \textbf{Netflix Movies and TV Shows}. Columns: \textit{date\_added}. \\ \url{https://www.kaggle.com/shivamb/netflix-shows}
        
        \item \textbf{Metacritic-Game Releases by Score}. Columns: \textit{Date}. \\ \url{https://www.kaggle.com/abhishekdataset/metacriticgame-releases-by-score}
    \end{itemize}

    \subsubsection{Numerical}
    \begin{itemize}
        \item \textbf{FIFA 19 complete player dataset}. Columns: \textit{LS, ST, RS, LW}. Model: \textit{GradientBoostingRegressor}. Task: \textit{Regression}. Target: \textit{Value}. \\
        \url{https://www.kaggle.com/karangadiya/fifa19}
        
        \item \textbf{HR Analytics: Job Change of Data Scientists}. Columns: \textit{company\_size}. \\ \url{https://www.kaggle.com/arashnic/hr-analytics-job-change-of-data-scientists?select=aug\_train.csv}
        
        \item \textbf{Students' Academic Performance Dataset}. Columns: \textit{StudentAbsenceDays}.
        \url{https://www.kaggle.com/aljarah/xAPI-Edu-Data}
    \end{itemize}
    
    \subsubsection{Sentence}
    \begin{itemize}
        \item \textbf{Wine Reviews}. Columns: \textit{description}. Model: \textit{GradientBoostingRegressor}. Task: \textit{Regression}. Target: \textit{points} \url{https://www.kaggle.com/zynicide/wine-reviews}
        
        \item \textbf{World Development Indicators}. Columns: \textit{SpecialNotes, SystemOfNationalAccounts}.  \url{https://www.kaggle.com/worldbank/world-development-indicators}
    \end{itemize}
    
    \subsubsection{URL}
    \begin{itemize}
        \item \textbf{FIFA 19 complete player dataset}. Columns: \textit{Photo, Flag, Club Logo}. Model: \textit{GradientBoostingRegressor}. Task: \textit{Regression}. \\ Target: \textit{Value}. \url{https://www.kaggle.com/karangadiya/fifa19}
        
        \item \textbf{Walmart Product Details 2020}. Columns: \textit{Product Url}. \\ \url{https://www.kaggle.com/promptcloud/walmart-product-details-2020}
    \end{itemize}
    
    \subsubsection{Zip code}
    \begin{itemize}
        \item \textbf{Data\_UK}. Columns: \textit{postal}. \url{https://www.kaggle.com/phool1804/data-uk}
        
        \item \textbf{OpenAddresses - Europe}. Columns: \textit{POSTCODE}. \\ \url{https://www.kaggle.com/openaddresses/openaddresses-europe?select=netherlands.csv}
        
        \item \textbf{OpenAddresses - North America (excluding U.S.)}. Columns: \textit{POSTCODE}. \url{https://www.kaggle.com/openaddresses/openaddresses-north-america-excluding-us?select=bermuda.csv}
        
        \item \textbf{House Price Data, England \& Wales, 2015 to 2019}. Columns: \textit{SS2 6ST}. Model: \textit{GradientBoostingRegressor}. Task: \textit{Regression}. Target: \textit{249995}. \url{https://www.kaggle.com/dmaso01dsta/house-price-data-england-wales-2015-to-2019}
    \end{itemize}

\subsection{Statistical type prediction and determining order in data}\label{app:datasets2}
    Some of the datasets listed below were used for both statistical type prediction and determining the order in ordinal data. The datasets without model, task, and target were only used to classify the statistical type based on the manually labeled ground truth of the columns. The datasets with the previously mentioned components were used for both statistical type prediction (with manual labeling of ordinality) and performance evaluation of FlairNLP (evaluation by running the associated task).
    
    \subsubsection{Nominal datasets}
    \begin{itemize}
        \item \textbf{[NeurIPS 2020] Data Science for COVID-19 (DS4C)}. Columns: \textit{province, city}. \url{https://www.kaggle.com/kimjihoo/coronavirusdataset?select=Case.csv}
        
        \item \textbf{[NeurIPS 2020] Data Science for COVID-19 (DS4C)}. Columns: \textit{type, gov\_policy}. \\ \url{https://www.kaggle.com/kimjihoo/coronavirusdataset?select=Policy.csv}
        
        \item \textbf{AB\_NYC\_2019}. Columns: \textit{name, host\_name, neighborhood\_group, neighborhood, room\_type}. \\ \url{https://www.kaggle.com/chadra/ab-nyc-2019}
        
        \item \textbf{Automobile Dataset}. Columns: \textit{make}. \\ \url{https://www.kaggle.com/toramky/automobile-dataset}
        
        \item \textbf{Craft Beers Dataset}. Columns: \textit{style}. \\ \url{https://www.kaggle.com/nickhould/craft-cans?select=beers.csv}
        
        \item \textbf{Craft Beers Dataset}. Columns: \textit{city, state}. \\ \url{https://www.kaggle.com/nickhould/craft-cans?select=breweries.csv}
        
        \item \textbf{FIFA 19 complete player dataset}. Columns: \textit{Nationality, Club}. \\ 
        \url{https://www.kaggle.com/karangadiya/fifa19}
        
        \item \textbf{FiveThirtyEight Comic Characters Dataset}. Columns: \textit{ALIGN, EYE, HAIR}.  \\
        \url{https://www.kaggle.com/fivethirtyeight/fivethirtyeight-comic-characters-dataset?select=dc-wikia-data.csv}
        
        \item \textbf{HR Analytics: Job Change of Data Scientists}. Columns: \textit{city, major\_discipline, company\_type}. \\ \url{https://www.kaggle.com/arashnic/hr-analytics-job-change-of-data-scientists?select=aug\_train.csv}
        
        \item \textbf{IBM HR Analytics Employee Attrition \& Performance}. Columns: \textit{Department, EducationField, JobRole}. Model: \textit{GradientBoostingRegressor}. Task: \textit{Regression}. Target: \textit{MonthlyIncome}. \\ \url{https://www.kaggle.com/pavansubhasht/ibm-hr-analytics-attrition-dataset}
        
        \item \textbf{Kickstarter Projects}. Columns: \textit{category, main\_category, currency, country}. \\ \url{https://www.kaggle.com/kemical/kickstarter-projects?select=ks-projects-201612.csv}
        
        \item \noindent \textbf{Mushroom Classification}. Columns: \textit{class, cap-shape, cap-surface, cap-color, bruises, odor, gill-attachment, gill-spacing, gill-size, gill-color, stalk-shape, stalk-root, stalk-surface-above-ring, stalk-surface-below-ring, stalk-color-above-ring, stalk-color-below-ring, veil-type, veil-color, ring-numer, ring-type, spore-print-color, population, habitat}. \\
        \url{https://www.kaggle.com/uciml/mushroom-classification}
        
        \item \textbf{Pokemon with stats}. Columns: \textit{Type 1, Type 2}. \\ \url{https://www.kaggle.com/abcsds/pokemon}
        
        \item \textbf{Ramen Ratings}. Columns: \textit{Brand, Variety, Style, Country}. \\ \url{https://www.kaggle.com/residentmario/ramen-ratings}
        
        \item \textbf{Stroke Prediction Dataset}. Columns: \textit{work\_type, smoking\_status}. \\ \url{https://www.kaggle.com/fedesoriano/stroke-prediction-dataset}
        
        \item \textbf{Students' Academic Performance Dataset}. Columns: \textit{PlaceOfBirth, GradeID, SectionID, Topic}. \\ \url{https://www.kaggle.com/aljarah/xAPI-Edu-Data}
        
        \item \textbf{Students Performance in Exams}. Columns: \textit{race/ethnicity}. \\ \url{https://www.kaggle.com/spscientist/students-performance-in-exams}
        
        \item \textbf{Wine Reviews}. Columns: \textit{country, province, region\_1, variety}.  Model: \textit{GradientBoostingRegressor}. Task: \textit{Regression}. Target: \textit{points}. \\
        \url{https://www.kaggle.com/zynicide/wine-reviews}
    \end{itemize}

    \subsubsection{Ordinal datasets}\label{app:datasets-order}
    \begin{itemize}
        \item \textbf{Amazon - Ratings (Beauty Products)}. Columns: \textit{Rating}. \\ \url{https://www.kaggle.com/skillsmuggler/amazon-ratings?select=ratings\_Beauty.csv}
        
        \item \textbf{Audiology (Original) Data Set}. Columns: \textit{air, ar\_c, ar\_u, bone, o\_ar\_c, o\_ar\_u, speech}. \\ \url{https://archive.ics.uci.edu/ml/datasets/Audiology+\%28Original\%29}
        
        \item \textbf{Basic Income Survey - 2016 European Dataset}. Columns: \textit{dem\_education \_level, awareness, vote, age\_group}. \\ \url{https://www.kaggle.com/daliaresearch/basic-income-survey-european-dataset}
        
        \item \textbf{Car Evaluation Data Set}. Columns: \textit{buying, maint, doors, persons, lug\_boot, safety, class value}. Model: \textit{GradientBoostingClassifier}. Task: \textit{Classification}. Target: \textit{Class Values}. \\ \url{https://archive.ics.uci.edu/ml/datasets/Car+Evaluation}
        
        \item \textbf{Earthquake Magnitude, Damage and Impact}. Columns: \textit{damage\_overall \_colapse, damage\_overall\_leaning, damage\_grade, technical\_solution\_proposed}. \\ \url{https://www.kaggle.com/arashnic/earthquake-magnitude-damage-and-impact?select=csv\_building\_damage\_assessment.csv}
        
        \item \textbf{Earthquake Magnitude, Damage and Impact}. Columns: \textit{education\_level \_household\_head}. \\ \url{https://www.kaggle.com/arashnic/earthquake-magnitude-damage-and-impact?select=csv\_household\_demographics.csv}
        
        \item \textbf{Hayes-Roth Data Set}. Columns: \textit{age, educational level, marital status}. \\
        \url{https://archive.ics.uci.edu/ml/datasets/Hayes-Roth}
        
        \item \textbf{Linux Gamers Survey, Q1 2016}. Columns: \textit{LinuxUserHowLong, DesktopLinuxGamerHowLong, HeavyGamer, LinuxExclusivity, LinuxGamingHabit-Change, LinuxGamingHabitFuture, LinuxGamingMachineShared, FolksAround-YouAwareLinux, LinuxGamesPurchaseFrequency, SatisfactionSteam, SatisfactionGOG, SatisfactionHB, DistroChangeFrequency, DistroImpactPerformance, HardwareUpgradeIntent, AwarenessBrandedSteamMachines, AwarenessSteamController, AwarenessSteamLink, SteamMachineConceptLike, SteamMachinesExpandLinuxDoubtful, SteamMachinesLaunchEvaluation, SteamMachinesAwarenessAlienware, SteamMachinesAwarenessZotac, SteamMachinesAwarenessSyber, SteamMachinesWantToBuy, MachinesMaximumPrice, MachinesDIYIntent, SteamControllerPurchaseIntent, SteamOSEverTried, SteamIHSUage, SteamLinkPurchaseIntent, WINEUsageVanilla, PlayOnLinux, Crossover, WINEEvaluation}. \\ \url{https://www.kaggle.com/sanqualis/linuxgamerssurvey}
        
        \item \textbf{Nursery Data Set}. Columns: \textit{parents, has\_nurs, form, housing, finance, social, health}. Model: \textit{GradientBoostingClassifier}. Task: \textit{Classification}. Target: \textit{Nursery}. \\ \url{https://archive.ics.uci.edu/ml/datasets/Nursery}
        
        \item \textbf{Solar Flare Data Set}. Columns: \textit{activity, evolution, previous\_24h\_flare\_acti-vity\_code, area}. Model: \textit{GradientBoostingRegressor}. Task: \textit{Regression}. Target: \textit{C-class, M-class, X-class}. \\ \url{https://archive.ics.uci.edu/ml/datasets/Solar+Flare}
        
        \item \textbf{Soybean (Large) Data Set}. Columns: \textit{precip, temp, crop-hist, area-damaged, severity, stem-cankers}. Model: \textit{GradientBoostingClassifier}. Task: \textit{Classification}. Target: \textit{class}. \\ \url{https://archive.ics.uci.edu/ml/datasets/Soybean+\%28Large\%29}
    \end{itemize}

\end{document}